# An Efficient Hierarchical Kriging Modeling Method for High-dimension Multi-fidelity Problems

Youwei He, Jinliang Luo[*]

School of Mechanical Engineering, University of South China, Hengyang 421001, China

**Abstract**: Multi-fidelity Kriging model is a promising technique in surrogate-based design as it can balance the model accuracy and cost of sample preparation by fusing low- and high-fidelity data. However, the cost for building a multi-fidelity Kriging model increases significantly with the increase of the problem dimension. To attack this issue, an efficient Hierarchical Kriging modeling method is proposed. In building the low-fidelity model, the maximal information coefficient is utilized to calculate the relative value of the hyperparameter. With this, the maximum likelihood estimation problem for determining the hyperparameters is transformed as a one-dimension optimization problem, which can be solved in an efficient manner and thus improve the modeling efficiency significantly. A local search is involved further to exploit the search space of hyperparameters to improve the model accuracy. The high-fidelity model is built in a similar manner with the hyperparameter of the low-fidelity model served as the relative value of the hyperparameter for high-fidelity model. The performance of the proposed method is compared with the conventional tuning strategy, by testing them over ten analytic problems and an engineering problem of modeling the isentropic efficiency of a compressor rotor. The empirical results demonstrate that the modeling time of the proposed method is reduced significantly without sacrificing the model accuracy. For the modeling of the isentropic efficiency of the compressor rotor, the cost saving associated with the proposed method is about 90% compared with the conventional strategy. Meanwhile, the proposed method achieves higher accuracy.

## 1. Introduction

Surrogate model, also known as metamodels or response surfaces, has been widely used in numerical optimization or uncertainty quantification for expensive engineering problems to replace the time-consuming simulation models, aiming to relieve the computational burden (HAN et al., 2020; Shu et al., 2019; Zhou et al., 2020). Various types of surrogate model have been developed, such as polynomial response surface models (Chatterjee et al., 2019; Hawchar et al., 2017), support vector regression models (Shi et al., 2020; Xie et al., 2018; Zhou et al., 2015), radial basis function models (Chen et al., 2022; Liu et al., 2022; Song et al., 2019), neural networks (Yegnanarayana, 1994) and Kriging models (J. Forrester et al., 2006). Among them, Kriging gains popularity as it can not only provides the predictions of the expensive models but also estimate the prediction errors. Based on Kriging, optimization methods for single- and multi-objective problems (Schonlau et al., 1998; Zhan et al., 2017; Zhan & Xing, 2020) and global sensitivity analysis methods (Cheng et al., 2020; Van Steenkiste et al., 2019) have been developed to solve practical problems for aerodynamic (He et al., 2020; Wang et al., 2018) or structural (Viana et al., 2014; Zhou et al., 2020) design applications.

Despite the continuous advance in Kriging-based modeling methods, the associated prohibitive computational cost of building sufficiently accurate Kriging for high-dimension applications remains an important challenge. Specifically, the cost of building the Kriging model for high-dimension problems is twofold. Firstly, to construct a sufficient accurate model, the number of sample data required will increase sharply. This will call for a large number of expensive simulations. Therefore, the cost of sample data preparation will be prohibitive for high-dimension problems. Secondly, with the increase of sample set, the cost for fitting the model will increase exponentially. For problems with plenty of parameters, the cost of model construction will be unacceptable. In extreme applications, the process of model tuning might even be more expensive than engineering simulations.

Incorporating cheap auxiliary information has been demonstrated to be a promising strategy to alleviate the computational burden of data preparation. Such cheap information usually refers to low-fidelity data or inexpensive gradients. In this paper, Kriging assisted with cheap low-fidelity data, termed as multi-fidelity Kriging, is concerned for its extensive application in many fields such as numerical design optimization (He et al., 2021, 2022; Lin et al., 2022) or modeling of complex simulation problem (Lin et al., 2021). Co-Kriging (Kennedy & O'Hagan, 2000), Hierarchical Kriging (Han & Görtz, 2012), generalized hierarchical Co-Kriging (Zhou et al., 2020) and etc. are typical multi-fidelity Kriging surrogate models. Among them, the Hierarchical Kriging (HK) model has gained popularity because its merit of being as accurate as Co-Kriging and as simple as the correction-based methods. For instance, the HK model has been adopted to develop the variable-fidelity Efficient Global Optimization method (HAN et al., 2020).

Though the cost of sample data preparation can be decreased by incorporating cheap low-fidelity data, the construction cost of multi-fidelity Kriging model remains inappropriate or even computational prohibitive for high-dimension problems. This is usually known as the curse of dimensionality for metamodels, for either single- or multi-fidelity model. To attack the curse for single-fidelity Kriging model, Toal et al. (Toal et al., 2008) suggested to use isotropic correlation function (i.e. the same hyperparameter for each variable) for high-dimension problems. Empirical comparison indicates that tuning a reduced set of hyperparameter might outperform an inaccurately tuned out but complete set of hyperparameters. Based on this observation, Zhao et al. (Zhao et al., 2020) developed an efficient Kriging modeling method based on maximal information coefficient. The relative magnitudes of hyperparameter are estimated by maximal information coefficient, or the importance of each variable is represented by the maximal information coefficient. Then this knowledge is utilized to reformulate the maximum likelihood estimation problem to reduce the dimensionality. Therefore, the modeling efficiency can be improved. It should be noted that if values of maximal information coefficient reflecting the importance of a variable is inconsistent with the reality, biased values of maximal information coefficient may even mislead the tuning process of hyperparameter. Instead of using maximal information coefficient, the distance correlation is adopted to represent the variable importance in (Fu et al., 2020). Furthermore, a high-dimension Kriging modeling method by utilizing the Partial Least Squares regression technique was developed in (Bouhlel et al., 2016b, 2016a). Partial Least Squares regression is adopted to reveal how inputs depend on responses and reduce the dimension. In this method, the number of hyperparameters is reduced to a maximum of four and the modeling time can be reduced remarkably. For multi-fidelity models based on Kriging, a multi-fidelity high dimensional model representation (MF-HDMR) is developed to efficiently approximate high dimensional problems (Cai et al., 2017). However, empirical thresholds are required to determine the linearity of the first-order component function and to test whether the second- or higher-order HDMR component exists or not. Moreover, the modeling time of the MF-HDMR over the test problems in the numerical experiments are not reported. Overall, it still remains a challenge to build a high-quality multi-fidelity Kriging model within a reasonable amount of computational effort for high-dimension problems.

To that end, an efficient HK modeling method for high-dimension multi-fidelity design problems is proposed. In building the low-fidelity Kriging model, the maximum likelihood estimation problem is transformed into a one-dimension problem with the help of the relative values of the hyperparameters estimated by the technique of sensitivity analysis. By solving the one-dimension problem, a rough estimation of the hyperparameter for the low-fidelity Kriging model can be obtained. To prevent the possible misleading of the biased values of the sensitivity indicator, a correction step is further involved to obtain a fine combination of the hyperparameters. Similar strategy is adopted in tuning the high-fidelity model. The difference is that the relative magnitudes of the hyperparameters of high-fidelity model is provided by the hyperparameters of the fine-tuned low-fidelity model. The performance of the efficient HK modeling method is illustrated by ten analytic test examples and one real-world engineering example. Comparison between the proposed strategy and existing approach in terms of both the modeling efficiency and accuracy are carried out.

The remainder of the paper is organized as follows. Section 2 briefs the theoretical background. Motivation and the proposed tuning strategy are detailed in Section 3. Numerical experiments over analytic test problems and the isentropic efficiency modeling of an axial flow compressor rotor are presented to demonstrate the effectiveness of the proposed method. Finally, conclusions and suggestions for future work are provided in Section 5.

## 2. Background

The objective of this paper is to develop an efficient HK modeling strategy for high-dimension multi-fidelity problems. For better understanding, the two-fidelity modeling problems is considered. The efficient modeling strategy can build an accurate enough model for the high-fidelity black-box function $y = f_{HF}(\mathbf{x})$ with the assistant of $y = f_{LF}(\mathbf{x})$ with lowest computational cost. $\mathbf{x} \in \mathbb{R}^d$ is the modeling variable with the number of variables being $d$. The low- and high-fidelity responses at a sampling site are usually obtained by simulations, like the computational fluid dynamic-based simulation.

### 2.1. Kriging

Kriging assumes that a random process exits in each sampling site. It includes the trend function and the random process. According to the trend function used, there exist simple, ordinary, and universal Kriging. For ordinary Kriging, the prediction formulation can be expressed as:

$$Y(\mathbf{x}) = \mu + Z(\mathbf{x}) \tag{1}$$

where $\mu$ represents the unknown constant; $Z(\mathbf{x})$ denotes a stationary random process with zero mean and process variance $\sigma^2$. The covariance of $Z(\mathbf{x})$ is formulated as:

$$\mathrm{Cov}(Z(\mathbf{x}), Z(\mathbf{x}')) = \sigma^2 R(\mathbf{x}, \mathbf{x}') \tag{2}$$

where $R(\mathbf{x}, \mathbf{x}')$ is the spatial correlation function depending on the Euclidean distance between two sites $\mathbf{x}$ and $\mathbf{x}'$. Various versions of correlation function can be utilized. In this paper, the Matérn 5/2 correlation function (Ulaganathan et al., 2015) is adopted, which is formulated as:

$$R(\mathbf{x},\mathbf{x}') = \left(1 + \sqrt{5}a + \frac{5a^2}{3}\right)\exp\left(-\sqrt{5}a\right) \tag{3}$$

where $a = \sqrt{\sum_{i=1}^{d}\theta_i|x_i - x_i'|^2}$; $d$ is the number of variables; $\boldsymbol{\theta} = [\theta_1, \theta_2, ..., \theta_d]$ are hyperparameters measuring the activity of each variable. $\boldsymbol{\theta}$ is determined in the model fitting process by solving the following maximum likelihood estimation problem:

$$\boldsymbol{\theta} = \arg\max\left(-\frac{m}{2}\ln\hat{\sigma}^2(\boldsymbol{\theta}) - \frac{1}{2}\ln|\mathbf{R}(\boldsymbol{\theta})|\right) \tag{4}$$

where $m$ is the number of samples; $\hat{\sigma}^2$ denotes the estimated value of $\sigma^2$; $\mathbf{R}$ is the correlation matrix. The likelihood function is often multimodal. Therefore, the evolutionary algorithms, such as genetic algorithm, are usually adopted to solve the optimization problem shown in (4). While, the evolutionary algorithms often need thousands fitness evaluations of the likelihood function. For high-dimension problems, the matrix inversion during the thousands evaluation of the likelihood function would result in prohibitive computational cost, which might even be more time-consuming than engineering simulations.

The Kriging prediction $\hat{y}(\mathbf{x})$ for the quality of interest at any unvisited point are expressed as:

$$\hat{y}(\mathbf{x}) = \mu^* + \mathbf{r}^\mathrm{T}\mathbf{R}^{-1}(\mathbf{y}_s - \mu^*\mathbf{1}) \tag{5}$$

where $\mu^*$ is obtained via generalized least-square estimation; $\mathbf{r}$ is the correlation vector between the unvisited point and the sampled points; $\mathbf{y}_s$ is the response vector containing the sample responses; $\mathbf{1}$ is a unit column vector.

## 2.2. Hierarchical Kriging

HK is one of the multi-fidelity Kriging models, which can fuse abundant low-fidelity sample data and a small set of high-fidelity data to obtain an approximation with high accuracy. Usually, the time cost for obtaining a low-fidelity sample is much cheaper than that of a high-fidelity sample. Therefore, the cost of sample data preparation can be reduced. In HK, the low-fidelity function is taken as the model trend for the high-fidelity model to avoid the calculation of the covariance matrix between low- and high-fidelity samples.

The construction of a HK model starts with tuning of the low-fidelity Kriging model based on the low-fidelity samples. Then, the low-fidelity Kriging is used as the model trend of the Kriging for the high-fidelity function, which is expressed as:

$$Y(\mathbf{x}) = \beta\hat{y}_{\mathrm{LF}}(\mathbf{x}) + Z(\mathbf{x}) \tag{6}$$

where $\beta$ is a scaling factor indicating the level of correlation between the low- and high-fidelity functions; $\hat{y}_{\mathrm{LF}}(\mathbf{x})$ denotes the prediction of low-fidelity Kriging; $Z(\mathbf{x})$ is the random process with zero mean and variance with the identical form as shown in (2). The parameters in the correlation function are determined in the model tuning procedure by maximizing the likelihood estimation problem.

The HK prediction is formulated as

$$\hat{y}(\mathbf{x}) = \beta^*\hat{y}_{\mathrm{LF}}(\mathbf{x}) + \mathbf{r}^\mathrm{T}\mathbf{R}^{-1}(\mathbf{y}_s - \beta^*\mathbf{F}) \tag{7}$$

where $\beta^* = (\mathbf{F}^\mathrm{T}\mathbf{R}^{-1}\mathbf{F})^{-1}\mathbf{F}^\mathrm{T}\mathbf{R}^{-1}\mathbf{y}_s$; $\mathbf{y}_s$ is the column vector containing the true responses of the high-fidelity sample; $\mathbf{F}$ represents the column vector of the predictions from the low-fidelity Kriging at the high-fidelity sample sites. More details are referred to (Han & Görtz, 2012).

For clarity, main steps for the conventional construction of a HK model are summarized below:

Step 1: Collect the low- and high-fidelity sample data $D_{\text{LF},n} = \{(\boldsymbol{x}_{\text{LF},i}, y_{\text{LF},i})\}_{i=1}^{n}$ and $D_{\text{HF},k} = \{(\boldsymbol{x}_{\text{HF},i}, y_{\text{HF},i})\}_{i=1}^{k}$;

Step 2: Determine the hyperparameters in the correlation function of the low-fidelity model by solving the following problem:

$$\boldsymbol{\theta}_{\text{LF}} = \arg\max\left(-\frac{n}{2}\ln\hat{\sigma}_{\text{LF}}^{2}(\boldsymbol{\theta}_{\text{LF}}) - \frac{1}{2}\ln|\mathbf{R}_{\text{LF}}(\boldsymbol{\theta}_{\text{LF}})|\right) \quad (8)$$

Step 3: Obtain the predictions from the low-fidelity Kriging at the high-fidelity sample sites via (5);

Step 4: Obtain the hyperparameters in the correlation function of the high-fidelity model by solving the following problem:

$$\boldsymbol{\theta}_{\text{HF}} = \arg\max\left(-\frac{k}{2}\ln\hat{\sigma}_{\text{HF}}^{2}(\boldsymbol{\theta}_{\text{HF}}) - \frac{1}{2}\ln|\mathbf{R}_{\text{HF}}(\boldsymbol{\theta}_{\text{HF}})|\right) \quad (9)$$

Step 5: Calculate the prediction at untested sites using (7).

In Step 2 and 4, the maximization problems are often solved by evolutionary algorithms. It will call for thousands or even more evaluation of likelihood function. However, for high-dimension problems, each calculation of likelihood function will be computational expensive. Therefore, the cost for model tuning of HK will be prohibitive for high-dimension problems. If the number of the parameters in the likelihood maximization problem can be reduced, the number of calls of the evaluation of likelihood function can be reduced, which means the modeling cost will be decreased significantly. Or if better initial values of the hyperparameters are available, local optimization method, which usually need less function evaluations, could be adopted to find better estimation of the hyperparameters with lower computational cost.

## 3. The efficient Hierarchical Kriging model

It has been demonstrated that the hyperparameter $\theta_i$ of Kriging model indicates the extent how the *i*th input variable influences the response (Forrester & Keane, 2009; Ulaganathan et al., 2015). In detail, a larger value of $\theta_i$ means that the *i*th variable has greater influence on the response. Meanwhile, sensitivity indicator in the field of sensitivity analysis can measure the importance of variables over the response (Shan & Wang, 2010). Therefore, if the relationship between sensitivity indicator and hyperparameter can be established, the dimensionality of the maximum likelihood estimation problems might be reduced to improve the modeling efficiency.

In HK, the hyperparameters of both the low- and high-fidelity model indicate the importance of the variable to the low- and high-fidelity response, respectively. Moreover, the low- and high-fidelity functions generally correlate well with each other. It is reasonable to believe that the hyperparameters of the low-fidelity model might measure the importance of the variable to the high-fidelity response as well. Or, there might be a linear or simple relationship between the hyperparameters of the low- and high-fidelity model. If such relationship can be revealed, it can be used to reduce the number of parameters in the likelihood maximization problems, or it may even serve as a good initial guess of the hyperparameter to narrow the search space of the hyperparameter so as to alleviate the computational burden of the model tuning procedure.

Above all, we would like to make use of two relationships to develop an efficient modeling strategy of HK. The first one is the relationship between the sensitivity indicator and hyperparameter of low-fidelity model. The other one is the relationship between the low- and high-fidelity hyperparameters. In this paper, the sensitivity indicator maximal information coefficient (MIC) is adopted, which is briefed firstly in this section. Then, an analytic example is introduced to illustrate the feasibility of the idea behind the proposed efficient modeling strategy of HK. After that. technique details and implementation are presented.

### 3.1. Maximal information coefficient

MIC is an sensitivity analysis method for identifying the variables with significant influence on the response (Reshef et al., 2011). It is an improved version of mutual information. The MIC of two variables $\mathbf{x}_l$ and $\mathbf{y}$ is expressed as:

$$\omega_l(\mathbf{x}_l, \mathbf{y}) = \max_{a,b<B} \frac{\mathrm{MI}(\mathbf{x}_l, \mathbf{y})}{\log_2(\min(a,b))} \tag{10}$$

where $\omega_l \in [0,1]$ is the MIC value; $a$ and $b$ are the number of rows and columns of gridding the scatterplot of data $\mathbf{x}_l$ and $\mathbf{y}$; $B$ is the upper bound of the grid size, $\mathrm{MI}(\mathbf{x}_l, \mathbf{y})$ denotes the mutual information between $\mathbf{x}_l$ and $\mathbf{y}$. In practice, the $\mathrm{MI}(\mathbf{x}_l, \mathbf{y})$ is estimated by the following formula:

$$\mathrm{MI}(\mathbf{x}_l, \mathbf{y}) = \sum_{i=1}^{n} \hat{p}(\mathbf{x}_l^{(i)}, \mathbf{y}^{(i)}) \log \frac{\hat{p}(\mathbf{x}_l^{(i)}, \mathbf{y}^{(i)})}{\hat{p}(\mathbf{x}_l^{(i)}) \hat{p}(\mathbf{y}^{(i)})} \tag{11}$$

where $\hat{p}(\mathbf{x}_l^{(i)})$ and $\hat{p}(\mathbf{y}^{(i)})$ denote the estimated probability density function, and $\hat{p}(\mathbf{x}_l^{(i)}, \mathbf{y}^{(i)})$ represents the estimated joint probability density function. Larger value of MIC implies greater influence of a variable on the response. Notably, the MIC does not assume any distribution of sample data and are easy to compute. In multi-fidelity modeling problems, the number of low-fidelity samples are usually much larger than that of the low-fidelity samples. Therefore, in the proposed method, the MICs between each variable and the response are estimated using low-fidelity data other than high-fidelity data, as a larger data set can result in more accurate identification of the influence of variable on the response via MIC.

### 3.2. An illustrative example

To clarify the motivation of the proposed method, an analytic function is utilized to illustrate the relationship between the MIC and hyperparameters of low-fidelity model as well as the relationship between hyperparameters of low- and high-fidelity model. The high- and low-fidelity function of the analytic problem from (Cai et al., 2017) is expressed as:

$$\begin{aligned} f_{\mathrm{HF}}(\mathbf{x}) &= x_1^2 + x_2^2 + x_1 x_2 - 14 x_1 - 16 x_2 + (x_3 - 10)^2 + 4(x_4 - 5)^2 + (x_5 - 3)^2 + 2(x_6 - 1)^2 \\ &\quad + 5 x_7^2 + 7(x_8 - 11)^2 + 2(x_9 - 10)^2 + (x_{10} - 7)^2 + 45 \\ f_{\mathrm{LF}}(\mathbf{x}) &= 0.8 f_{\mathrm{HF}} - \sum_{i=1}^{10} x_i + 100 \\ x_i &\in [-10, 11], i = 1, 2, \ldots, 10 \end{aligned} \tag{12}$$

To begin with, 100 and 50 sample data are collected for the low- and high-fidelity model, respectively. Before the calculation of MIC and the construction of the models, low- and high-fidelity sample data are centered to have zero mean. The HK model is built with the collected sample data following the procedure described in Section 2.2. Specifically, the genetic algorithm is adopted to solve the likelihood maximization problem. The population size is set as 40, and the maximum number of function valuations is 5000. The fractions of crossover and migration are set as 0.8 and 0.2, respectively. To prevent the influence of the random procedure in the genetic algorithm, the HK model is built on the identical sample set by 20 times. 5000 high-fidelity test data is adopted to measure the accuracy of the built model. The most accurate model is screened out and the hyperparameters are recorded. Finally, the MIC values, hyperparameters of the low- and high-fidelity model are plotted in Fig. 1.

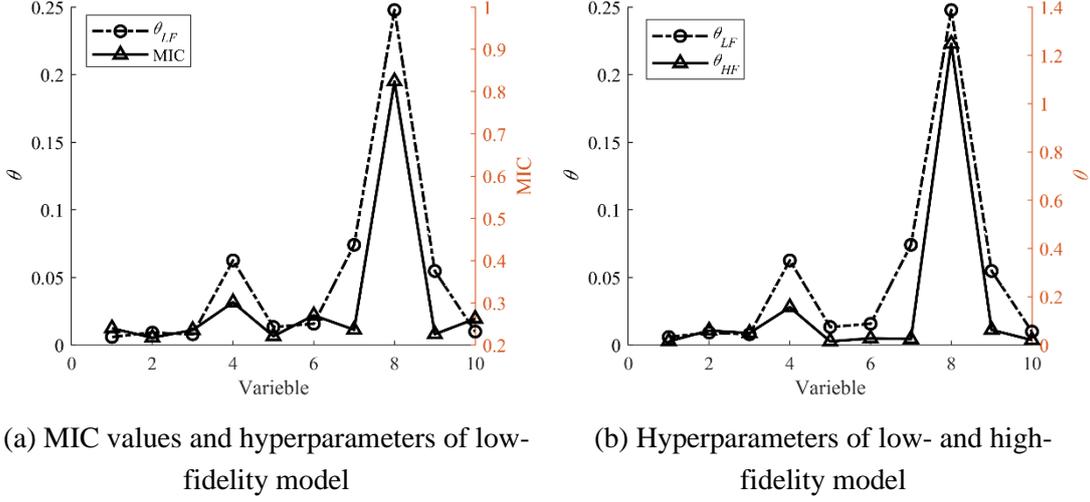

(a) MIC values and hyperparameters of low-fidelity model

(b) Hyperparameters of low- and high-fidelity model

**Fig. 1.** Hyperparameters and MIC values for the illustration function

As shown in Fig. 1(a), the trends of the MIC values and the tuned hyperparameters are quite similar. Moreover, it can be noted that both the MIC and tuned hyperparameters can identify the importance of each variable. From the expression of the current problem, it can be observed that $x_8$ has the largest coefficient and should be the most influential variable of this problem. Such observation can also be concluded from Fig. 1(a) with the MIC and hyperparameter. The MIC values and the hyperparameters of $x_1, x_2, x_3$ are small, indicating that those three variables have less influence on the response. This can also be confirmed from the function expression. As the MIC values and tuned hyperparameter has similar trends but different magnitude, it is possible to assume that the hyperparameters are proportional to the MIC values. Or, the following linear relationship can be established:

$$\mathbf{\theta}_{LF} = \lambda \mathbf{\omega} \tag{13}$$

where $\lambda \in \mathbb{R}^+$ is the scale factor between the MIC values $\mathbf{\omega}$ and the hyperparameters of the low-fidelity model $\mathbf{\theta}_{LF}$. As shown in Fig. 1(b), the hyperparameters of the low- and high-fidelity model share nearly identical trends but with different magnitude. It is natural to believe that the hyperparameters of the high-fidelity model are proportional to the hyperparameters of the low-fidelity model. Such observation can be expressed as follows:

$$\mathbf{\theta}_{HF} = \chi \mathbf{\theta}_{LF} \tag{14}$$

where $\chi \in \mathbb{R}^+$ is the scale factor between the hyperparameters of the low- and high-fidelity model $\boldsymbol{\theta}_{LF}$ and $\boldsymbol{\theta}_{HF}$. This illustrative example exemplified the inner relationship between the sensitivity indicator and hyperparameters as well as the connection between the low- and high-fidelity model hyperparameters. Then the question left is how to make full use of those relationships to improve the modeling efficiency of the HK model, which is depicted in next subsection.

### 3.3. Proposed construction strategy

With above observations, the hyperparameter estimation problem for the low-fidelity model shown in (8) can be reformulated as follows:

$$\boldsymbol{\theta}_{LF} = \arg\max\left(-\frac{n}{2}\ln \hat{\sigma}_{LF}^2(\boldsymbol{\theta}_{LF}) - \frac{1}{2}\ln\left|\mathbf{R}_{LF}(\boldsymbol{\theta}_{LF})\right|\right) \quad (15)$$
$$s.t.\ \boldsymbol{\theta}_{LF} = \lambda\boldsymbol{\omega}$$

In practice, $\boldsymbol{\theta}_{LF}$ is obtained with a two-step strategy. Firstly, the above equality constrained problem is reformulated into an unconstrained problem by inserting the equality relationship between MIC and $\boldsymbol{\theta}_{LF}$ to determine $\lambda$:

$$\lambda = \arg\max\left(-\frac{n}{2}\ln \hat{\sigma}_{LF}^2(\lambda\boldsymbol{\omega}) - \frac{1}{2}\ln\left|\mathbf{R}_{LF}(\lambda\boldsymbol{\omega})\right|\right) \quad (16)$$

Then $\boldsymbol{\theta}_{LF} = \lambda\boldsymbol{\omega}$ is utilized to calculate $\boldsymbol{\theta}_{LF}$. Compared with the $d$-dimension optimization problem (8), the hyperparameter estimation problem is now a one-dimension problem. It can be, of course, solved in a more efficient manner than the original one. The number of likelihood evaluations can be reduced significantly, thus improving the modeling efficiency.

While, such strategy has a drawback obviously. The hyperparameters of each variable are tied together with the scale factor λ artificially. During the tuning process, the hyperparameters cannot change independently, which might sacrifice the model accuracy. Moreover, the importance of a variable estimated by MIC might be inconsistent with the reality. The biased MIC values may mislead the tuning process, degrading the effectiveness of the proposed strategy. Therefore, a local search is further involved with solving the problem (8) by starting from the already obtained $\boldsymbol{\theta}_{LF}$. The local search allows the independent changes of each hyperparameter. This is much useful to improve the accuracy of the low-fidelity model based on our preliminary investigation. Low-fidelity model with high accuracy can further improve the performance of the multi-fidelity model. It is one of the key points to ensure the effectiveness of the proposed efficient HK model.

The tuning process for the high-fidelity model is similar to that of the low-fidelity model. The main difference is the utilization of the connection between the hyperparameters of low- and high-fidelity model. In detail, the following one-dimension problems is solved firstly to obtain an estimation of the scale factor between the hyperparameters of the low- and high-fidelity model:

$$\chi = \arg\max\left(-\frac{k}{2}\ln\hat{\sigma}_{HF}^2(\chi\boldsymbol{\theta}_{LF}) - \frac{1}{2}\ln|\mathbf{R}_{HF}(\chi\boldsymbol{\theta}_{LF})|\right) \quad (17)$$

Then, $\boldsymbol{\theta}_{HF}$ is calculated by $\boldsymbol{\theta}_{HF} = \chi\boldsymbol{\theta}_{LF}$. After that, a local search starting from the already obtained $\boldsymbol{\theta}_{HF}$ is followed to improve the model accuracy by allowing the independent change of each hyperparameter.

For clarity, the main steps of the proposed efficient HK modeling method for multi-fidelity high-dimension problems are summarized below:

Step 1: Collect the low- and high-fidelity sample data $D_{LF,n} = \{(\boldsymbol{x}_{LF,i}, y_{LF,i})\}_{i=1}^n$ and $D_{HF,k} = \{(\boldsymbol{x}_{HF,i}, y_{HF,i})\}_{i=1}^k$;

Step 2: Calculate the values of MIC $\boldsymbol{\omega}$ by using the low-fidelity data $D_{LF,n}$;

Step 3: Determine the scale factor $\lambda$ by solving the problem in (16);

Step 4: Obtain the $\boldsymbol{\theta}_{LF}$ via the relationship $\boldsymbol{\theta}_{LF} = \lambda\boldsymbol{\omega}$;

Step 5: Solve the problem in (8) by a local optimizer starting from the hyperparameters obtained in Step 4 to obtain a better estimation of $\boldsymbol{\theta}_{LF}$;

Step 6: Obtain the predictions from the low-fidelity Kriging at the high-fidelity sample sites via (5);

Step 7: Determine the scale factor $\chi$ by solving the problem in (17);

Step 8: Obtain the $\boldsymbol{\theta}_{HF}$ via the relationship $\boldsymbol{\theta}_{HF} = \chi\boldsymbol{\theta}_{LF}$;

Step 9: Solve the problem in (9) by a local optimizer starting from the hyperparameters obtained in Step 8 to obtain a better estimation of $\boldsymbol{\theta}_{HF}$;

Step 10: Calculate the prediction at untested sites using (7).

The proposed HK modeling method is implemented based on a DACE toolbox (Lophaven et al., 2002). In the generation of the sample sites, the Latin hypercube sampling is adopted. The minepy package (Albanese et al., 2013) is adopted to calculate the MIC values with default settings. The one-dimension optimization problems in the model fitting process are solved via the Matlab's fminbnd optimizer. For the local optimizer, Matlab's fmincon function is utilized. Options for those optimizers and details of the implementation can be found in the source code, which is available at https://github.com/Youwei-He/HDHK. To verify the implementation, the Forrester function (Forrester et al., 2007) is employed. The high- and low-fidelity function are given by:

$$\begin{aligned} f_{HF}(\mathbf{x}) &= (6x^2 - 2)\sin(12x - 4) \\ f_{LF}(\mathbf{x}) &= 0.5 f_{HF} + 10(x - 0.5) - 5 \\ x &\in [0,1] \end{aligned} \quad (18)$$

The sampling sites for the high- and low-fidelity data are $x_{HF} = \{0, 0.4, 0.6, 1\}$ and

$x_{\text{LF}} = \{0, 0.1429, 0.2857, 0.4286, 0.5714, 0.7143, 0.8571, 1\}$, respectively. Two HK models are built with the conventional and proposed strategy. Figure 2 compares the low-fidelity and high-fidelity predictions with the true functions. The predictions from the model tuned by the proposed strategy are labeled with HD as the method is developed for high-dimension problems. Overall, the low- or high-fidelity predictions from either tuning strategy agree well with the true functions. The predictions from the two modeling strategies almost overlap with each other.

The $\beta^*$ estimated with the conventional and the proposed strategy is 1.8769 and 1.8772, respectively, which both are close to the true value of 2. Those observations verify the implementation. The time for the conventional and proposed tuning strategy are 0.1216s and 0.0243s, respectively. This indicates that, though the strategy is developed for tuning the high-dimension HK model efficiently, it can also improve the modeling efficiency on this one-dimension problem. To quantify the performance of the proposed method, numerical experiments are carried out and presented in next section.

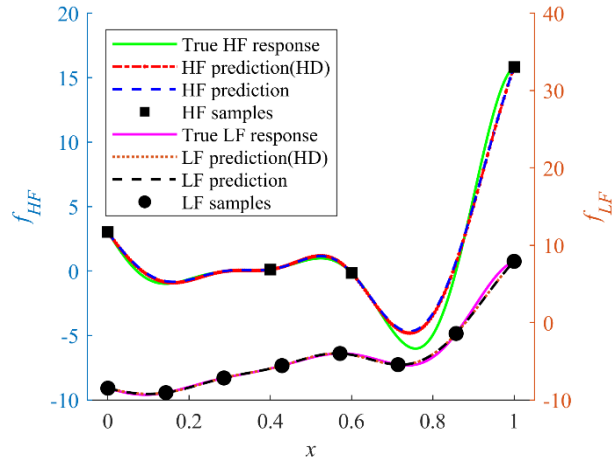

**Fig. 2.** HK predictions over the Forrester function

## 4. Experimental study

In this section, the performance of the proposed method is tested and compared with conventional tuning strategy. For simplicity, the HK employing the conventional tuning strategy and the proposed high-dimension modeling method is shorted as HKC and HKHD, respectively.

### 4.1. Numerical examples

The expressions of the adopted analytic test problems (Cai et al., 2017) are summarized in Table 1. The number of modeling variables ranges from 2 to 50.

**Table 1.** Numerical test functions

| No. | Function | Design Space |
|---|---|---|
| 1 | $f_{\text{HF}}(\mathbf{x}) = 4x_1^2 - 2.1x_1^4 + \frac{1}{3}x_1^6 + x_1 x_2 - 4x_2^2 + 4x_2^4$ <br> $f_{\text{LF}}(\mathbf{x}) = f_{\text{HF}}(0.7\mathbf{x}) + x_1 x_2 - 65$ | $x_i \in [-2, 2]$ <br> $i = 1, 2$ |
| 2 | $f_{HF}(\mathbf{x}) = \left(x_2 - 1.275\left(\frac{x_1}{\pi}\right)^2 - 5\frac{x_1}{\pi} - 6\right)^2 + 10\left(1 - \frac{0.125}{\pi}\right)\cos(x_1)$ <br> $f_{LF}(\mathbf{x}) = 0.8 f_{HF}(\mathbf{x}) - 2.5 x_2 - 30$ | $x_1 \in [-5, 10]$ <br> $x_2 \in [0, 15]$ |
| 3 | $f_{HF}(\mathbf{x}) = 100(x_1^2 - x_2)^2 + a_1^2 + a_3^2 + 90(x_3^2 - x_4) + 10.1(a_2^2 + a_4^2) + 19.8 a_2 a_4$ <br> $a_i = x_i - 1, i = 1, 2, 3, 4$ <br> $f_{LF}(\mathbf{x}) = 90(x_1^2 - x_2)^2 + a_1^2 + a_3^2 + 50(x_3^2 - x_4) + 5(a_2^2 + a_4^2) + 10 a_2 a_4$ <br> $a_i = 0.9 x_i - 1, i = 1, 3;\quad a_i = 0.5 x_i - 1, i = 2, 4$ | $x_i \in [-4, 4]$ <br> $i = 1, \ldots, 4$ |
| 4 | $f_{HF}(\mathbf{x}) = \sum_{i=1}^{10} \exp(x_i)\left[A(i) + x_i - \ln\left(\sum_{k=1}^{10} \exp(x_k)\right)\right]$ <br> $A = [-6.089, -17.164, -34.054, -5.914, -24.721, -14.986, -24.100,$ <br> $-10.708, -26.662, -22.179]$ <br> $f_{LF}(\mathbf{x}) = \sum_{i=1}^{10} \exp(x_i)\left[B(i) + x_i - \ln\left(\sum_{k=1}^{10} \exp(x_k)\right)\right]$ <br> $B = [-5, -10, -30, -5, -25, -15, -20, -10, -25, -20]$ | $x_i \in [-5, 5]$ <br> $i = 1, \ldots, 10$ |
| 5 | $f_{HF}(\mathbf{x}) = \sum_{i=1}^{9}\left((x_{i+1}^2 - x_i)^2 + (x_i - 1)^2\right)$ <br> $f_{LF}(\mathbf{x}) = \sum_{i=1}^{9}\left(0.9 x_{i+1}^4 + 2.2 x_i^2 - 1.8 x_i x_{i+1}^2 + 0.5\right)$ | $x_i \in [-3, 3]$ <br> $i = 1, \ldots, 10$ |
| 6 | $f_{HF}(\mathbf{x}) = x_1^2 + x_2^2 + x_1 x_2 - 14 x_1 - 16 x_2 + (x_3 - 10)^2 + 4(x_4 - 5)^2 + (x_5 - 3)^2$ <br> $+ 2(x_6 - 1)^2 + 5 x_7^2 + 7(x_8 - 11)^2 + 2(x_9 - 10)^2 + (x_{10} - 7)^2 + 45$ <br> $f_{LF}(\mathbf{x}) = 0.8 f_{HF} - \sum_{i=1}^{10} x_i + 100$ | $x_i \in [-10, 11]$ <br> $i = 1, \ldots, 10$ |
| 7 | $f_{HF}(\mathbf{x}) = (x_1 - 1)^2 + \sum_{i=2}^{16} i(2 x_i^2 - x_{i-1})^2$ <br> $f_{LF}(\mathbf{x}) = 0.9 f_{HF}(\mathbf{x}) + 10$ | $x_i \in [-5, 5]$ <br> $i = 1, \ldots, 16$ |
| 8 | $f_{HF}(\mathbf{x}) = (x_1 - 1)^2 + \sum_{i=2}^{30} i(2 x_i^2 - x_{i-1})^2$ <br> $f_{LF}(\mathbf{x}) = 0.8 f_{HF}(\mathbf{x}) - \sum_{i=1}^{29} 0.4 x_i x_{i+1} - 50$ | $x_i \in [-3, 3]$ <br> $i = 1, \ldots, 30$ |
| 9 | $f_{HF}(\mathbf{x}) = \sum_{i=1}^{50} i(x_i^2 + x_i^4)$ <br> $f_{LF}(\mathbf{x}) = 0.8 f_{HF}(\mathbf{x}) - \sum_{i=1}^{50}(i x_i^2 / 10 + x_i) - 25$ | $x_i \in [-2, 4]$ <br> $i = 1, \ldots, 50$ |

For HKC, the likelihood maximization problems are solved by Genetic Algorithm. The population size is set as $4d$, and the maximum generation is set as 125. Therefore, the maximum number of likelihood function evaluation is $500d$. The fractions of crossover and migration are set as 0.8 and 0.2, respectively. The search space of the hyperparameter $\boldsymbol{\theta}$ is $[10^{-4}, 10^2]^d$. The maximum number of function evaluation for the fminbnd optimizer to solve the one-dimension problem in (16) and (17) is set as 500. Search interval for the scale factor $\lambda$ and $\chi$ is set as $[10^{-4}, 10^2]$ based on preliminary test. For the fmincon optimizer utilized to do the local search, function evaluations as many as 500 times are allowed. The number of low- and high-fidelity samples is set as $10d$ and $5d$, respectively.

To test the accuracy of the models, $200d$ (maximum 5000) validation points are generated by Latin hypercude sampling. Two global accuracy metrics, the coefficient of determination $R^2$ and root mean square error (RMSE), and a local accuracy metric, the maximum absolute error (MAE), are utilized to evaluate the model accuracy:

$$R^2 = 1 - \frac{\sum_{i=1}^{N}(y_i - \hat{y}_i)^2}{\sum_{i=1}^{N}(y_i - \bar{y})^2} \tag{19}$$

$$MSE = \sqrt{\sum_{i=1}^{N}\frac{(y_i - \hat{y}_i)^2}{N}} \tag{20}$$

$$MAE = \max(|y_i - \hat{y}_i|) \tag{21}$$

where $N$ denotes the number of validation points; $y_i$ and $\hat{y}_i$ are the true and predicted high-fidelity response of the $i$th validation point, respectively; $\bar{y}$ is the mean of the true response of validation points. Notably, only the high-fidelity prediction is involved in the accuracy comparison, as the high-fidelity response is usually the quantity of interest in practical applications. A closer value to 1 of $R^2$, indicates the better global accuracy of the model. Smaller values of the RMSE and MAE means better accuracy. Moreover, the training time are recorded to measure the modeling efficiency of the two strategies. Each analytic problem is modelled 10 times to obtain the mean and standard deviation (STD) results of those metrics. The experiment is conducted as a PC with Intel Xeon CPU E5-2666 v3 @ 2.90GHz and 64GB RAM.

**4.2. Results and discussion**

Table 2 summarizes the statistic results of the modeling time and accuracy metrics. Boxplots are utilized to better visualize the test results over representative problems in Fig. 3-7. It can be noted that the modeling time of HKHD is much shorter than that of the HKC in all the test functions. Generally, the modeling time of HKHD method is 1/7~1/10 to that of the HKC method. For 10-D No. 4 function, the mean modeling time of HKHD is 0.589s, while it is 5.474s averagely for HKC method. This indicates that the proposed method can save more than 95% time for constructing the HK model. For the 50-D No.9 test function, the modeling time of HKHD is 304.0s in average. It saved 85.2% time compared with the conventional tuning strategy, which need 2055.2s to construct the model. The time-saving would be much meaningful for applications, e.g., surrogate-based optimization, that the surrogate model needs to be tuned frequently.

In terms of the model accuracy, the HKHD is more accurate that the HKC over all numerical test functions. For the 4-D No.3 function, the $R^2$ of the HKHD and HKC are 0.990 and 0.659, respectively. The RMSE of the HKHD and HKC are 630.190 and 3591.430, respectively. This means that the global accuracy of HKHD is better that that of the HKC. As for the local accuracy, the HKHD outperforms the HKC as the MAE are 3909.955 and 19877.880, respectively. The $R^2$ of the HKHD over the 10-D No. 6 function is as high as 0.995, while it is 0.642 averagely of the HKC method. The RMSE and MAE values indicate that the global and local accuracy by HKHD method is 85.7% and 80.1% higher than that of the HKC. In the 50-D No.9 function, the HKHD achieved the performance of $R^2$ being 0.745, which is significantly higher than that of the HKC ($R^2$ being 0.305 averagely). The global and local accuracy indicated by the RMSE and MAE increased 42.9% and 41.7%, respectively. It should be mentioned that the performance of the model built following the conventional strategy might be improved by allowing more likelihood function evaluations. While, this might increase the modeling time significantly but the gain of accuracy might be unworthy, especially for applications which needs adaptive update of model. As a conclusion of the empirical experiments, the proposed HKHD method can build more accurate model than the conventional strategy within significant shorter time.

Table 2. Comparison of the metric statistic results

| No. | | Time(s) | | $R^2$ | | RMSE | | MAE | |
|---|---|---|---|---|---|---|---|---|---|
| | | HKC | HKHD | HKC | HKHD | HKC | HKHD | HKC | HKHD |
| 1 | Mean | 0.275 | 0.039 | 0.244 | 0.653 | 10.1 | 6.1 | 40.0 | 27.0 |
| | STD | 0.033 | 0.005 | 0.581 | 0.530 | 4.4 | 4.3 | 11.2 | 11.8 |
| 2 | Mean | 0.247 | 0.035 | 0.730 | 0.985 | 102.7 | 23.5 | 530.7 | 133.7 |
| | STD | 0.031 | 0.004 | 0.146 | 0.011 | 33.2 | 9.3 | 184.7 | 75.6 |
| 3 | Mean | 0.844 | 0.089 | 0.659 | 0.990 | 3591.4 | 630.2 | 19877.9 | 3910.0 |
| | STD | 0.107 | 0.018 | 0.288 | 0.008 | 1739.6 | 246.7 | 7594.0 | 1774.9 |
| 4 | Mean | 5.474 | 0.589 | 0.472 | 0.617 | 1328.9 | 1151.1 | 6471.3 | 5073.5 |
| | STD | 0.253 | 0.088 | 0.202 | 0.033 | 252.2 | 49.5 | 1317.1 | 841.4 |
| 5 | Mean | 5.564 | 0.401 | 0.288 | 0.456 | 77.7 | 66.6 | 360.7 | 300.5 |
| | STD | 0.051 | 0.179 | 0.239 | 0.304 | 14.3 | 18.2 | 58.8 | 79.5 |
| 6 | Mean | 5.553 | 0.520 | 0.642 | 0.995 | 477.1 | 68.1 | 1983.9 | 394.6 |
| | STD | 0.054 | 0.098 | 0.410 | 0.002 | 377.8 | 12.2 | 1107.7 | 91.2 |
| 7 | Mean | 28.593 | 4.309 | 0.459 | 0.705 | 19353.5 | 14497.4 | 100586.7 | 68426.1 |
| | STD | 6.281 | 0.604 | 0.199 | 0.041 | 3468.8 | 991.0 | 18000.1 | 8388.5 |
| 8 | Mean | 273.679 | 43.437 | 0.436 | 0.704 | 6639.1 | 4825.3 | 33752.2 | 22112.9 |
| | STD | 39.280 | 4.272 | 0.093 | 0.023 | 549.0 | 184.7 | 2797.1 | 3298.7 |
| 9 | Mean | 2055.243 | 304.094 | 0.305 | 0.745 | 11065.9 | 6309.7 | 44894.0 | 26156.2 |
| | STD | 12.531 | 102.526 | 0.118 | 0.249 | 963.1 | 2352.9 | 2352.8 | 8681.8 |

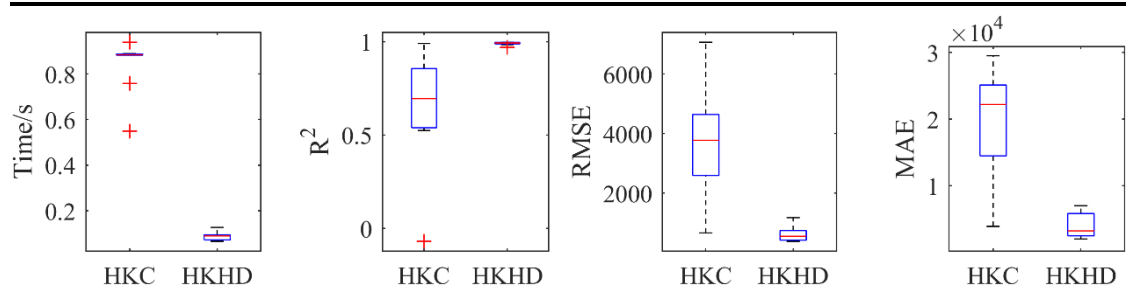

Fig. 3 Boxplots for the 4-D No. 3 function

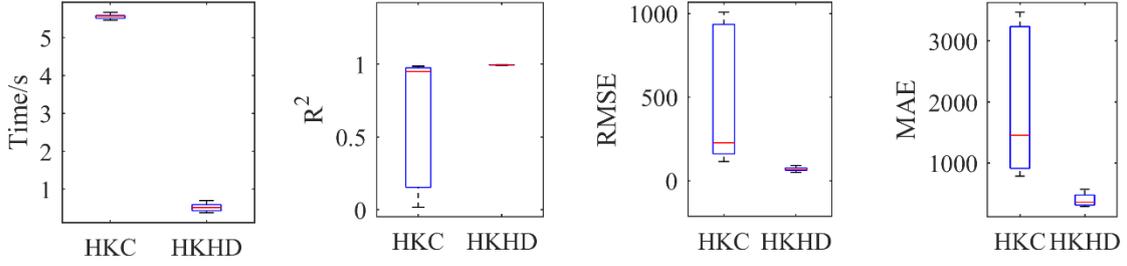

**Fig. 4** Boxplots for the 10-D No. 6 function

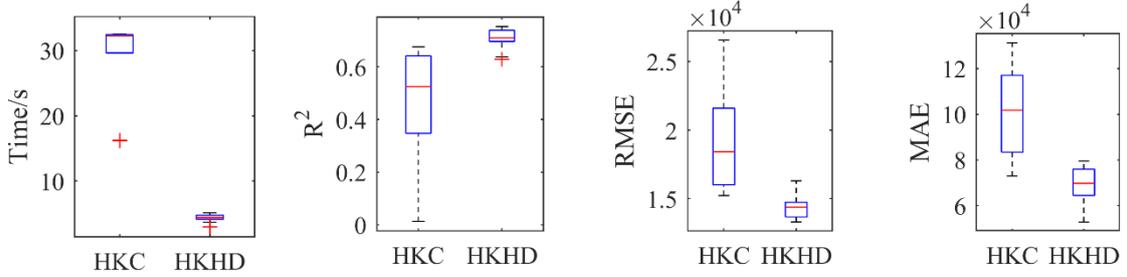

**Fig. 5** Boxplots for the 16-D No. 7 function

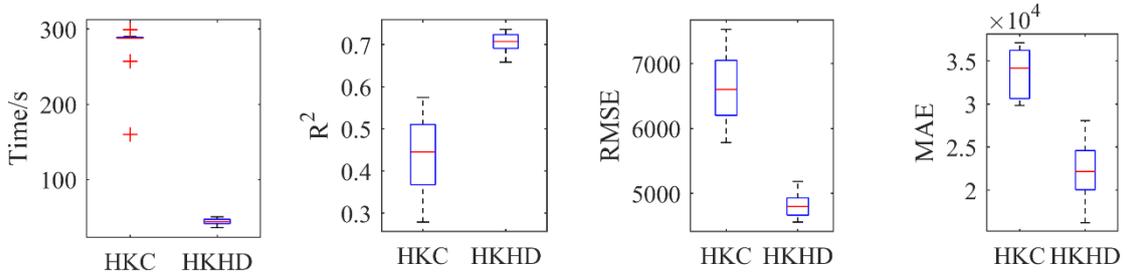

**Fig. 6** Boxplots for the 30-D No. 8 function

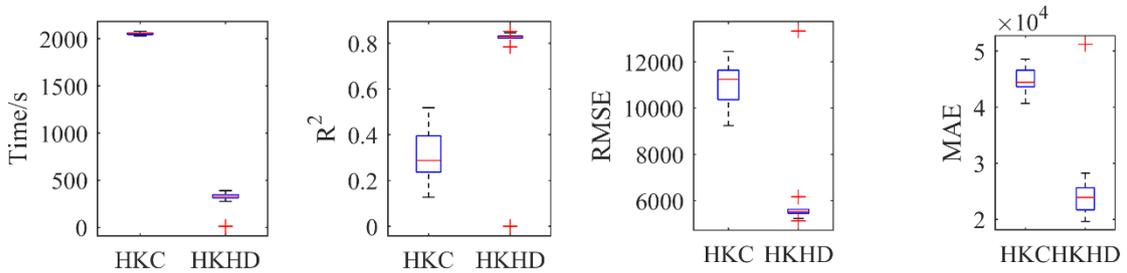

**Fig. 7** Boxplots for the 50-D No. 9 function

To investigate the effect the sample size on the modeling performance, additional two groups of experiment are conducted. One group of experiment start with the number of low- and high-fidelity samples being 8$d$ and 4$d$, respectively. The number of low- and high-fidelity samples is set as 12$d$ and 6$d$, respectively, in the other group of experiment. Table 3 presents the statistic results of the performance metrics by HKHD with different sample sizes. The metric values of the sample size being 10$d$+5$d$ taken from Table 2 are also included for better comparison. It can be noted that the modeling time increases with the increase of the sample size. For example, it needs 1.295s, 4.309s and 9.830s to finish the model construction of the No. 7 function with the sample size being 8$d$+4$d$, 10$d$+5$d$, and 12$d$+6$d$, respectively. For the 50-D No. 9 test problem, the modeling time increased from 304.0s to 3386.9s averagely, a 1003% increase, as the sample size expanded from 10$d$+5$d$ to 12$d$+6$d$. In terms of the model accuracy, it improves as more samples are adopted. For instance, the RMSE decreased from 6309.7 to 6165.3, 2.3% improvement as the sample size expanded from 10$d$+5$d$ to 12$d$+6$d$. While, for the sample size increased from 10$d$+5$d$ to 12$d$+6$d$, the global accuracy indicated by RMSE values (11031.2 and 6309.7) improved 42.8%. Those observations indicate that a moderate sample size would achieve a balance between the modeling efficiency and accuracy.

**Table 3.** Statistic results of the performance metrics by HKHD with different sample sizes

| Metric | Sample size | No. 3 Mean | No. 3 STD | No. 6 Mean | No. 6 STD | No. 7 Mean | No. 7 STD | No. 9 Mean | No. 9 STD |
|---|---|---|---|---|---|---|---|---|---|
| Time/s | 8$d$+4$d$ | 0.079 | 0.021 | 0.465 | 0.099 | 1.295 | 0.584 | 72.131 | 98.346 |
|  | 10$d$+5$d$ | 0.089 | 0.018 | 0.520 | 0.098 | 4.309 | 0.604 | 304.094 | 102.526 |
|  | 12$d$+6$d$ | 0.118 | 0.033 | 0.547 | 0.100 | 9.830 | 6.984 | 3386.934 | 1109.306 |
| RMSE | 8$d$+4$d$ | 813.7 | 350.7 | 87.7 | 32.1 | 18080.9 | 4248.7 | 11031.2 | 3560.8 |
|  | 10$d$+5$d$ | 630.2 | 246.7 | 68.1 | 12.2 | 14497.4 | 991.0 | 6309.7 | 2352.9 |
|  | 12$d$+6$d$ | 521.7 | 201.3 | 33.5 | 5.3 | 14391.7 | 4165.7 | 6165.3 | 2398.4 |
| MAE | 8$d$+4$d$ | 5166.5 | 2445.7 | 379.4 | 143.9 | 78327.7 | 22710.2 | 44964.6 | 14781.4 |
|  | 10$d$+5$d$ | 3910.0 | 1774.9 | 394.6 | 91.2 | 68426.1 | 8388.5 | 26156.2 | 8681.8 |
|  | 12$d$+6$d$ | 2718.1 | 981.8 | 156.6 | 24.3 | 69269.3 | 21991.6 | 24802.6 | 11418.1 |

Table 4 presents the statistic results of the performance metrics by HKC with different sample sizes. The modeling time increases with sample size. For example, it needs 108.9s to build the HK model with the sample size being 12$d$+5$d$ on the 16-D No.7 test problem. Meanwhile, the construction time is 28.5s and 11.7s averagely by using a sample with size of 10$d$+5$d$ and 8$d$+4$d$, respectively. While, with the increase of the sample size, the model accuracy is not always improved. The mean RMSE values of HKC on No. 9 problem are 11768.9, 11065.9, and 11115.2, respectively. The comparison of the performance metrics with the change of sample size between the HKHD and HKC are presented in Fig. 8-11. It can be noted that the HKHD outperforms HKC over those test problems with various sample size in terms of both modeling efficiency and accuracy.

**Table 4.** Statistic results of the performance metrics by HKC with different sample sizes

| Metric | Sample size | No. 3 Mean | No. 3 STD | No. 6 Mean | No. 6 STD | No. 7 Mean | No. 7 STD | No. 9 Mean | No. 9 STD |
|---|---|---|---|---|---|---|---|---|---|
| Time/s | 8$d$+4$d$ | 0.758 | 0.048 | 4.341 | 0.080 | 11.784 | 0.707 | 1283.576 | 39.593 |
|  | 10$d$+5$d$ | 0.844 | 0.107 | 5.553 | 0.054 | 28.593 | 6.281 | 2055.243 | 12.531 |
|  | 12$d$+6$d$ | 1.143 | 0.137 | 10.040 | 0.508 | 108.974 | 22.316 | 24942.824 | 241.766 |
| RMSE | 8$d$+4$d$ | 3289.1 | 1985.6 | 356.1 | 238.1 | 18721.6 | 3153.8 | 11768.9 | 500.7 |
|  | 10$d$+5$d$ | 3591.4 | 1739.6 | 477.1 | 377.8 | 19353.5 | 3468.8 | 11065.9 | 963.1 |
|  | 12$d$+6$d$ | 2089.8 | 1734.9 | 260.8 | 253.3 | 19427.9 | 3585.2 | 11115.2 | 567.8 |
| MAE | 8$d$+4$d$ | 16290.2 | 6844.3 | 1533.7 | 630.6 | 87603.9 | 13576.6 | 50396.9 | 1819.9 |
|  | 10$d$+5$d$ | 19877.9 | 7594.0 | 1983.9 | 1107.7 | 100586.7 | 18000.1 | 44894.0 | 2352.8 |
|  | 12$d$+6$d$ | 13361.1 | 8547.7 | 1225.0 | 744.1 | 100452.0 | 9498.5 | 50534.2 | 2838.8 |

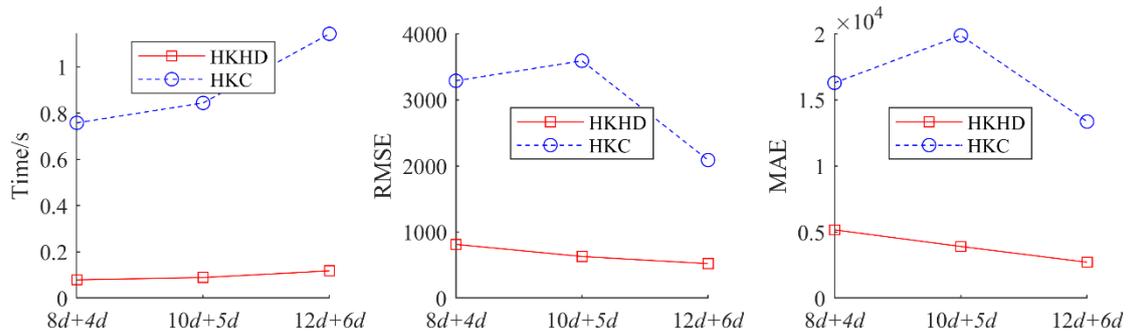
**Fig. 8** Evolution of the performance metrics with change of sample size on No. 3 problem

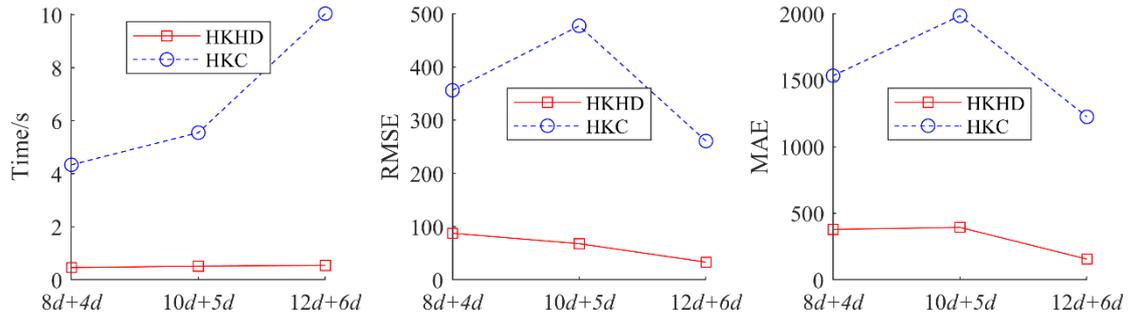
**Fig. 9** Evolution of the performance metrics with change of sample size on No. 6 problem

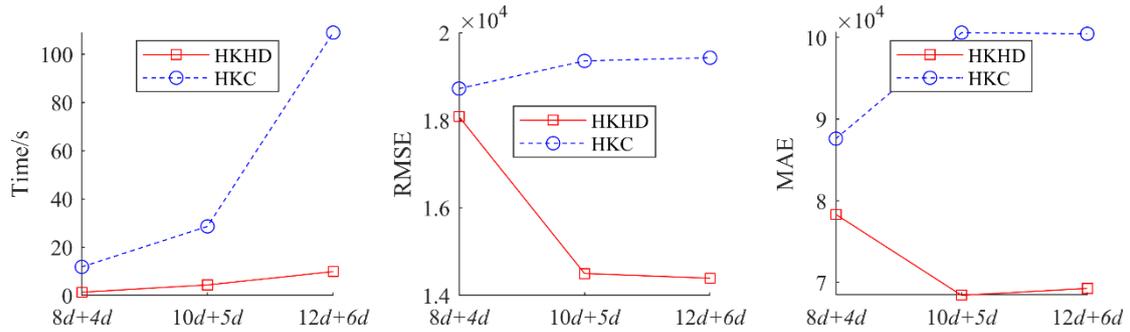
**Fig. 10** Evolution of the performance metrics with change of sample size on No. 7 problem

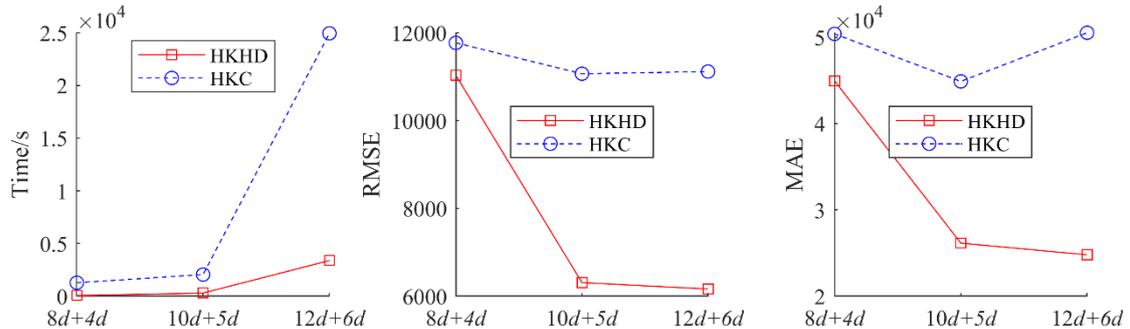
**Fig. 11** Evolution of the performance metrics with change of sample size on No. 9 problem

### 4.3. Engineering example

In addition to those analytic problems, an engineering problem of modeling the isentropic efficiency of the axial compressor rotor Rotor37 is covered to further demonstrate the effectiveness of the developed efficient modeling method. Rotor37 is an isolated axial-flow compressor wheel designed and experimentally analyzed at the NASA (Reid & Moore, 1978). The main geometric and design specifications of Rotor 37 are summarized in Table 5. Fig. 12 illustrates is 3-D view.

**Table 5.** Main geometric parameters and design specifications of Rotor37

| Parameter | Value |
|---|---|
| Mass flow rate | 20.2kg/s |
| Design pressure ratio | 2.106 |
| Design adiabatic efficiency | 87.6% |
| Number of blades | 36 |
| Design RPM | 17188 |
| Tip speed | 454m/s |
| Inlet Hub/Tip ratio | 0.70 |
| Aspect ratio | 1.19 |
| Tip solidity | 1.3 |

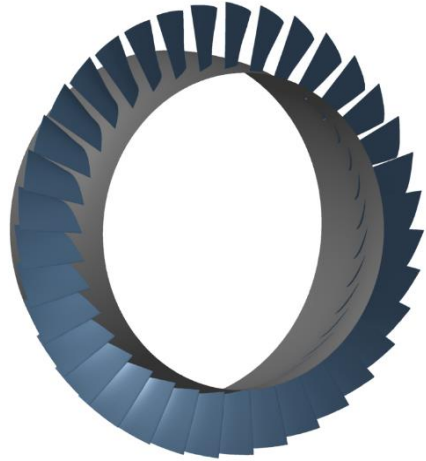

**Fig. 12** 3-D view of the Rotor37

The problem is to build a model to predict the isentropic efficiency of Rotor 37 with variation of the blade geometry:

$$f = \eta_c(\mathbf{x}) \quad (22)$$

with

$$\eta_c = \frac{h_{2s} - h_1}{h_{2r} - h_1} \quad (23)$$

where $h_1$ denotes the specific enthalpy of the air at the rotor inlet, $h_{2s}$ and $h_{2r}$ represent the specific enthalpy of the gas at the outlet of the rotor for isentropic and real compression process, respectively; $\mathbf{x}$ is the parameters determining the blade shape. $h_1$, $h_{2s}$, and $h_{2r}$ are obtained from the result of the computational fluid dynamic (CFD) simulation. In this problem, the Rotor37 blade is constructed with three blade sections and a stacking law. Each section is composed by adding the thickness of the suction and pressure side to the camber line. Camber line and thickness of pressure and suction side are parameterized by Bezier curve. For each section, nine parameters are used to determine the profile shape as illustrated in Fig. 13(a). In detail, $\beta_1$ and $\beta_2$ are the inlet and out blade angle, respectively; $\alpha$ and $\gamma$ denote the trailing wedge angle and camber angle. $t_{p1}$, $t_{p2}$, $t_{s1}$, $t_{s2}$, and $t_{s3}$ are the control point of the thickness distribution of the pressure and suction side. The line goes through the center of gravity of each section is the stacking line. As shown in Fig. 13(b), the profile at the blade mid and tip are allowed to in the axial and circumferential direction, usually known as sweep and lean of the stacking line. More details about the shape parameterization can be found in (NUMECA, 2021). As a result, the blade shape is governed by 31 parameters. The NUMECA/AutoBlade is utilized to generate the file describing the blade shape for the grid generation.

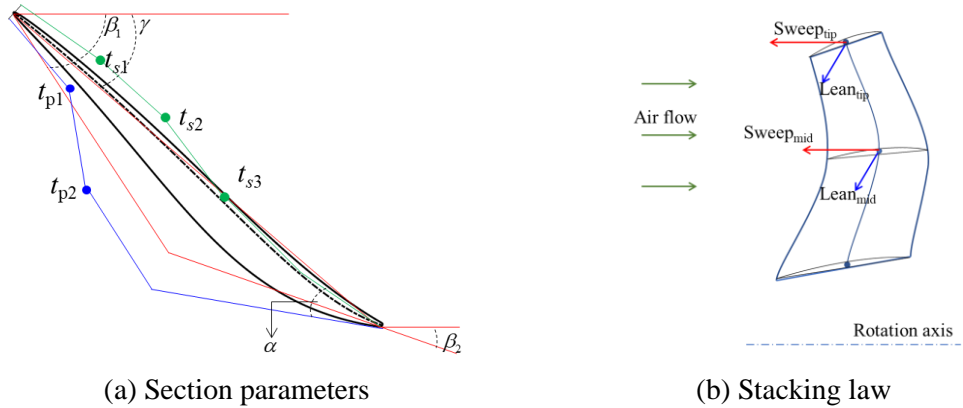

(a) Section parameters     (b) Stacking law

**Fig. 13** Geometric parameters for the parametric representation of the blade

Fine and coarse grids are used in the high- and low-fidelity simulation, respectively. Multi-block structured mesh is generated where the O4H topology is used around the blade and additional H-blocks are placed in the upstream/downstream of the blade. Refinements are conducted in the near walls to capture the boundary layer flow characteristics. Fig. 14 presents the grids for the low- and high-fidelity simulations for the baseline geometry, with the number of cells being 312077 and 799185, respectively.

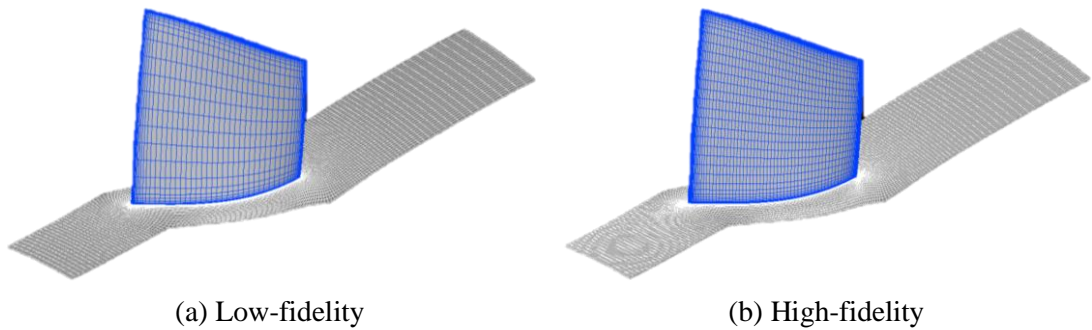

(a) Low-fidelity     (b) High-fidelity

**Fig. 14** Grid for low- and high-fidelity simulation

The CFD simulation solving the Reynolds time-averaged Navier-Stokes equations enclosed with the Spalart-Allmaras turbulence model are conducted in NUMECA to determine the isentropic efficiency. At the inlet, the total temperature and total pressure of the axial inlet flow are applied. While the static pressure is specified at outlet boundary. No-slip and adiabatic conditions at solid surfaces are applied and periodicity condition is applied at lateral sides of the computational domain to facilitate single blade passage simulation. Those boundary conditions are kept unchanged among all the simulations over different blades. Simulations stop if the global residual decreased to $10^{-5}$. The low- and high-fidelity simulation finished within about 5min and 12min, respectively. The isentropic efficiency of the Rotor37 obtained from the low- and high-fidelity simulations is 85.41% and 85.09%, respectively.

200 high-fidelity samples and 400 low-fidelity samples are generated by the Latin hypercube sampling procedure. Corresponding simulations are conducted to obtain the responses. 164 out of the 200 high-fidelity simulations ended smoothly. Meanwhile, for the 400 low-fidelity simulations, 296 simulations obtained the responses successfully. Rest of the low- and high-fidelity simulations failed. From the log of the computation, it is found that the simulation failures are resulted from bad geometry, ill mesh, and weak convergence of the CFD solver. The HKC and HKHD models are built based on this set of low- and high-fidelity data. To measure the performance of the models, 350 samples are generated by Latin hypercube sampling and simulated with the high-fidelity simulation. In turn, 285 computations are successful.

The performance metrics of those model predictions are listed in Table 6. The HKHD method spent 16.8s to build the HK model. This is a 90% saving of the modeling time compared with HKC, which spent 227.4s to tune the model. As for the accuracy, HKHD outperformed the HKC in terms of the $R^2$, RMSE and MAE metric. The $R^2$ is 0.975 and 0.907 for the HKHD and HKC, respectively, indicating the HKHD model is more accurate in the global view. For the local accuracy, HKHD is also superior to the HKC as the MAE values for those two models are 0.0295 and 0.0463. respectively. Moreover, comparison results of the simulation validation data and the predictions are illustrated in Fig. 15. It can be intuitively found that the HKHD method performs better than the HKC strategy. Overall, the proposed modeling strategy can build more accurate model with in significantly shorter time. This demonstrates the effectiveness of the proposed method on practical engineering problem.

**Table 6.** Metric values on the engineering problem

| Method | Time/s | $R^2$ | RMSE | MAE |
| --- | --- | --- | --- | --- |
| HKC | 227.4 | 0.907 | 0.0064 | 0.0463 |
| HKHD | 16.8 | 0.975 | 0.0033 | 0.0295 |

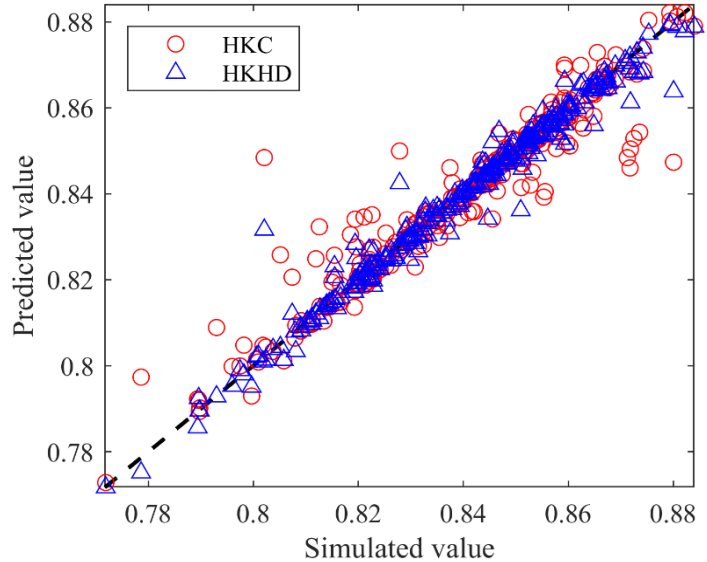

**Fig. 15** Comparisons of the simulated values and the predictions

## 5. Conclusions

In this paper, an efficient HK modeling method is developed for improving the modeling efficiency over high-dimension multi-fidelity problems. The relative magnitudes of hyperparameters are estimated by maximal information coefficients or the hyperparameters of lower fidelity model. Then the high-dimension maximum likelihood estimation problem is reformulated into a one-dimension problem to improve the modeling efficiency. Local correction search is added to further exploit the search space of the hyperparameters. To demonstrate the effectiveness and efficiency, ten numerical cases and one engineering modeling problem are tested. For the numerical examples, the proposed method only needs 1/7~1/10 time of the compared conventional strategy and can achieve higher accuracy. With the increase of the sample size, the modeling efficiency of the proposed method decreases and the model accuracy improves. For the conventional tuning strategy, the modeling efficiency decreases with the expansion of the sample set but the model accuracy is not always improved. As for the prediction of the isentropic efficiency of Rotor37, the cost saving associated with the proposed approach is about 90% compared with the conventional tuning strategy, and the proposed approach even achieves higher accuracy. Currently, the proposed method is illustrated for two-fidelity problems. We believe that extending the efficient modeling method to multi-fidelity problems would be straightforward. Moreover, optimization method based on the proposed efficient HK method will be pursued in the near future.